\documentclass[runningheads,a4paper,10pt]{llncs}
\usepackage[utf8]{inputenc}

\usepackage{amssymb}
\usepackage{tikz-qtree}
\usepackage{amsmath}
\usepackage{graphicx}
\usepackage{mathtools}
\usepackage{algorithmicx}
\usepackage{algorithm,algpseudocode}
\algnewcommand\algorithmicinput{\textbf{INPUT:}}
\algnewcommand\INPUT{\item[\algorithmicinput]}
\algnewcommand\algorithmicoutput{\textbf{OUTPUT:}}
\algnewcommand\OUTPUT{\item[\algorithmicoutput]}

\newcommand{\BigO}[1]{\ensuremath{\operatorname{O}\bigl(#1\bigr)}}
\DeclarePairedDelimiter{\ceil}{\lceil}{\rceil}
\DeclarePairedDelimiter{\floor}{\lfloor}{\rfloor}

\usepackage{url}
\urldef{\mailsa}\path|{aghaee, ghadiri}@ce.sharif.edu|
\urldef{\mailsb}\path|soleymani@sharif.edu|

\newcommand{\keywords}[1]{\par\addvspace\baselineskip
\noindent\keywordname\enspace\ignorespaces#1}

\begin{document}

\mainmatter  

\title{Active Distance-Based Clustering using K-medoids}

\titlerunning{Active Distance-Based Clustering using K-medoids}

\author{Amin Aghaee\footnotemark[1]\and Mehrdad Ghadiri\footnotemark[1]\and  Mahdieh Soleymani Baghshah}


\footnotetext[1]{Equally Contributing Authors}

\authorrunning{A. Aghaee, M. Ghadiri, and M. Soleymani Baghshah}

\institute{School of Computer Engineering, Sharif University of Technology, Iran\\
\mailsa\\
\mailsb\\
}

\toctitle{Lecture Notes in Computer Science}
\tocauthor{Authors' Instructions}
\maketitle

\begin{abstract}
k-medoids algorithm is a partitional, centroid-based clustering algorithm which uses pairwise distances of data points and tries to directly decompose the dataset with $n$ points into a set of $k$ disjoint clusters. However, k-medoids itself requires all distances between data points that are not so easy to get in many applications. In this paper, we introduce a new method which requires only a small proportion of the whole set of distances and makes an effort to estimate an upper-bound for unknown distances using the inquired ones. This algorithm makes use of the triangle inequality to calculate an upper-bound estimation of the unknown distances. Our method is built upon a recursive approach to cluster objects and to choose some points actively from each bunch of data and acquire the distances between these prominent points from oracle. Experimental results show that the proposed method using only a small subset of the distances can find proper clustering on many real-world and synthetic datasets.
\end{abstract}

\section{Introduction}
\label{sec:introduction}
As the production of data is expanding at an astonishing rate and the era of big data is coming, organizing data via assigning items into groups is inevitable. Data clustering algorithms try to find clusters of objects in such a way that the objects in the same cluster are more similar to each other than to those in other clusters. Nowadays clustering algorithms are widely used in data mining tasks.\par

There are different categorization for clustering algorithms, e.g., these algorithms can be categorized into Density-based, Centroid-based and Distribution-based methods. In centroid-based clustering methods, each cluster is shown by a central object. This object which can be a member of the dataset denotes a prototype of the whole cluster.  When these algorithms are appointed to find \textit{K} clusters, they usually find \textit{K} central objects and assign each element to the nearest centroid. As they go on, they attempt to decrease the energy and total error of clusters by finding better central elements. \textit{K-medoids} and \textit{K-means} are the two most popular centroid-based algorithms. Although, they both partition the data into groups such that the sum of the squared distances of the data points to the nearest center to them is minimized, they have different assumptions about centroids. Indeed, the k-medoids algorithm chooses the centroids only from the data points and so these centroids are members of the whole dataset while k-means algorithm can select the centroids from the whole input space. \par

In~\cite{albez2013kmd}, some usages and applications of k-medoids algorithm are discussed. According to this study, in resource allocation problems, when a company wants to open some branches in a city, in such a way that the average distance from each residential block to the closest branch is intended to be minimized, the k-medoids algorithm is a proper option. Additionally, in mobile computing context, it is an issue to save communication cost when devices need to choose super-nodes among each other, which should have minimum average distances to all devices and k-medoids can solve this problem. Furthermore, as reported by~\cite{albez2013kmd}, medoid queries also arise in the sensor networks and many other fields. \par

Active learning is a machine learning field that have bee attended specially in the last decade. Until now, many active methods for supervised learning that intend to select more informative samples to be labeled have been proposed. Active unsupervised methods have also been received attention recently. In an unsupervised learning manner, finding the similarities or distances of samples from each other may be difficult or infeasible. For example, the sequence similarity of proteins \cite{konstantin2012} or similarity of face images \cite{arijit2014} which needs to be obtained from human as an oracle, may be difficult to be responded. The active version of some of the well known clustering algorithms have been recently presented in \cite{Mai2013actdbscan,wang2010}. \par

In this paper, we propose the active k-medoid algorithm that inquires a subset of pairwise distances to find the clustering of data. We use a bottom-up approach to find more informative subset of the distances to be inquired. Extensive experiments on several data sets show that our algorithm usually needs a few percentage of the pairwise distances to cluster data properly.\par

In the rest of this paper, we first discuss about the works that have been done in the field of active clustering in Section \ref{sec:related}. In Section \ref{sec:method}, we introduce our algorithm. The result of experiments on different datasets have been presented in Section \ref{sec:experiments}. At last, we discuss about some aspects of our algorithm and conclude the paper in Section \ref{sec:conclusion}.

\section{Related Work}
\label{sec:related}
Active learning is a machine learning paradigm that endeavors to do learning with asking labels of a few number of samples which are more important in the final result of learning. Indeed, most of supervised learning algorithms need a large amount of labeled samples and gathering these labeled samples may need unreasonable amount of time and effort. Thus, active learning tries to ask labels for more important samples where important samples may be interpreted as most informative ones, most uncertain ones, or the ones that have a large effect in the results \cite{settles2010active}.
The active clustering problem has been recently received much attention. Until now, the active version of some well known clustering methods has been proposed in \cite{Mai2013actdbscan,wang2010}.
In the active clustering problem, a query is a pair of data whose 
similarity must be determined. The purpose of the active learning approach is reducing the number of required queries via active selection of them 
instead of random selection \cite{tong2001}. \par

The existing active clustering methods can be categorized into constraint-based and distance-based ones \cite{viet-vu2012} . 
In the most of the constraint-based methods, must-link and cannot-link constraints on pairs of data points indicating these pairs must be in the same cluster or different clusters are inquired.
Some constraint-based methods for active clustering have been proposed in \cite{viet-vu2012,xiong2014,grira2008,wang2010,arijit2014,basu2004,wagstaff2001}. 
In distance-based methods, the response to a query on a pair of data points is the distance 
of that pair according to an objective function. 
Distance-based methods for active clustering have been recently attended in \cite{Mai2013actdbscan,eriksson2011,krishnamurthy2012,Shamir2011,Wauthier2012,konstantin2012}. \par

In \cite{Mai2013actdbscan}, an algorithm for active DBSCAN clustering is presented. In this algorithm, the distances that have not been queried are estimated with a lower bound. A score indicating the amount and the probability of changes in the estimated distances by asking a query is used to select queries. Moreover, an updating technique is introduced in \cite{Mai2013actdbscan} that update clustering after a query. \par

In \cite{Shamir2011,Wauthier2012}, distance-based algorithms are presented for active spectral clustering in which a perturbation theory approach is used to select queries. 
A constraint-based algorithm has also been presented in \cite{wang2010} for active spectral clustering that uses an approach based on maximum expected error reduction to select queries. \par

An active clustering method for k-median clustering has also been proposed in \cite{konstantin2012}. This method selects some points as the landmarks and ask the distances between these landmarks and all the other data points as queries. Finally, k-median clustering is done using these distances.

\section{Proposed Method}
\label{sec:method}
\newcommand{\ABS}[1]{\left | #1 \right |}

In this section, we introduce the proposed active k-medoids clustering. We assume that our algorithm intends to partition $n$ samples into $k$ different clusters. As mentioned above, many clustering algorithms such as K-medoids, PAM \cite{Kaufman1990PAM}, and some other distance-based methods, calculate an \(n\times n\) distance matrix at first and perform the clustering algorithm on this distance matrix. We show the distance matrix by $D$ where 
$d_{ij}$ denotes the distance between the $i$th sample and the $j$th one. \par

We introduce a method to estimate unknown distances during an active clustering process. In a metric space, a satisfying and eminent upper-bound estimation for any distance metric can be obtained by the triangle inequality. 
For example, when we know the exact distances between $d_{ax},d_{xy} $ and $d_{yb}$, we can determine the upper-bound estimation for $d_{ab}$ as:
\begin{equation}
d_{ab} \leqslant d_{ax} + d_{xy} + d_{yb}
\end{equation}

We find an upper-bound estimation of the distances using the triangle inequality and the known distances asked from an oracle already.
Therefore, we have
\begin{equation}
\forall i,j, 1\leqslant i,j \leqslant n : D(i,j)\leqslant D_{e}(i,j)
\end{equation} \par
where $D_{e}$ shows the estimated distances.

First, upper-bound estimations for all distances are infinity and we update these distances by asking some of them and make better estimations for the other unknown distances using the triangle inequality and new distances taken from the oracle. The update will be done by replacing exact values for the asked distances and getting the minimum of the old and the new upper-bound estimation for unknown distances. By asking the landmark distances, we intend to take a better estimation of distances required for the k-medoids algorithm.

\begin{figure}[ht!]
\includegraphics[width=120mm]{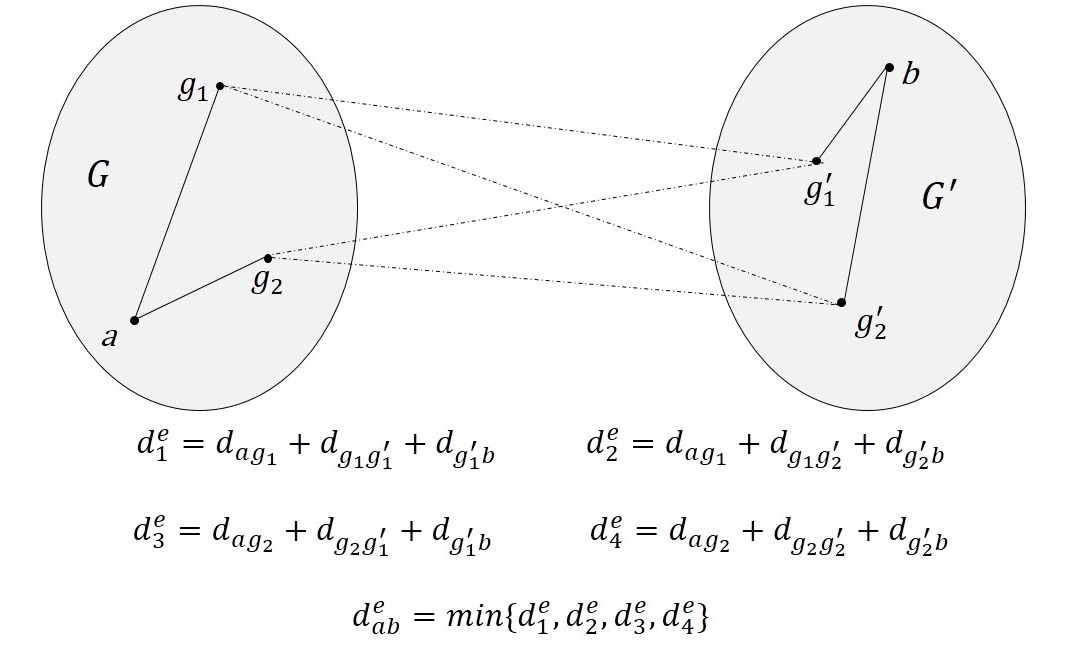}
\caption{Upper-bound distance estimation between $a,b$. Superscript $e$ determines that the distance is an estimation.\label{fig:estimation}}
\end{figure}

Consider some data points which are partitioned to $m$ groups where the distances within each group are known or estimated. However, the distances between data points from different groups are unknown. Our goal is to estimate these unknown distances instead of asking them. In such situation, we can choose $t$ finite points from each group and ask the distances between these $mt$ points between different groups and estimate the other distances using these asked distances. The number of these distances is ${m \choose 2}t^{2}$.
Figure \ref{fig:estimation} gives an intuition about this distance estimation method. 
We want to estimate the distance between $a$ and $b$ and the distances between those points that are connected by solid lines and dotted lines are known.

Such estimation algorithm can be done in $\BigO{n^2t^2}$, when $n$ is the number of data points. If we have $t \ll n$, the time complexity of the algorithm will be $\BigO{n^2}$.

Based on the estimation method used for unknown distances, we present an active k-medoids algorithm. 
The approach is based on partitioning the data points into some groups by a hierarchical manner. 
In the other words, we partition data points into $b$ groups and partition each of these groups to $b$ groups and so on until we get to a threshold like $t_h$ for the number of data points in each partition.
In this level, we ask all the distances within each group among all its data points and choose $t$ points from each group (to ask their distances) and using the explained algorithm for estimating distances in a bottom-up approach until we get to the highest level.
After that, we cluster the data using the k-medoids algorithm on the estimated distances.
According to these explanation, it seems that choosing $t$ points in each group is a critical step and choosing a bad point can lead to unfavourable estimations.
For this purpose, consider a group of data points like $G$ which its inner distances are known or estimated. 
In order to choosing $t$ points, we perform k-medoids algorithm on $G$ and find clusters and medoids for this group.
Then, we choose medoids and $s$ random points from each cluster as the points whose distances are needed to be asked.
Therefore, the number of the chosen data points in $G$ will be $t = k(s+1)$.
It is obvious that a greater $s$ will lead to more accurate estimations. \par

Algorithm \ref{alg:red} present the pseudo code of the proposed method.
This function clusters $n$ data points into $k$ different categories and return clusters of data.
Here, $b$ shows the branching factor which is used to partition data points to $b$ different groups with the same size.
The partitioning algorithm will perform for each group recursively. 
There is also a threshold $t_h$ which clarify the minimum size of a group of data points. Clearly, $t_h \geq k(s+1)$ since we need to choose at least $k(s+1)$ points in each group. It is also noteworthy that if $n \leq 2t_h$, we need to ask all distance pairs since these data points cannot be partitioned. \par

\begin{algorithm}
\caption{Active k-medoids}\label{alg:red}
\begin{algorithmic}[1]
\INPUT $D_e,n,k,b,t_h$ \Comment{distance estimation, \#data, \#clusters, branch factor, threshold}
\OUTPUT $C_1,\ldots,C_k$ 
\Procedure{ActiveKmedoids}{$D_e,n,k,b,t_h$}
\If{ $n \leq 2t_h$ } 
	\State Update $D_e$ by querying all distances 
	\State $C_1,\ldots,C_k \gets kmedoids(D_e,k)$ \Comment{kmedoids function is a regular k-medoids}
	\State \textbf{return}
\EndIf
\State Partition data to $b$ equal size groups, like $G_1,\ldots,G_b$
\For{\texttt{$i$ from $1$ to $b$}}
	\State $T_1,\ldots,T_k \gets ActiveKmedoids(D_e(G_i),|G_i|,k,b,t_h)$ \Comment{$D_e(G_i)$ is the part of the estimated distance matrix corresponding to $G_i$} 
	\State $G^c_i \gets$ medoids of $T_1,\ldots,T_k$ and $s$ random points from each of them.
\EndFor
\State Update $D_e$ by querying distances between all those pairs that one of them is in $G^c_i$ and the other is in $G^c_j$.
\State Update $D_e$ by the triangle inequality and new inquired distances.
\State $C_1,\ldots,C_k \gets kmedoids(D_e,k)$
\EndProcedure
\end{algorithmic}
\end{algorithm}

\begin{figure}[h]
\tikzset{edge from parent/.style={draw, edge from parent path=
    {(\tikzparentnode) -- (\tikzchildnode)}}
   ,level distance={1.5in},sibling distance={.2in}}
\begin{tikzpicture}[every tree node/.style={draw,rectangle,minimum width=1.1in,
    minimum height=.65in,align=center},scale=0.9]
	\Tree [.\node (7) {$n=1600$\\$data(1:1600)$\\$Query(list(3),list(6))$\\$Estimate D$};  
				\edge node [auto=right] {$data(1:800)$};
				[.\node (3) {$n=800$\\$Query(list(1),list(2))$\\$Estimate D$};
					\edge node [auto=right] {$data(1:400)$};
					[.\node (1) {$n=400$\\$Query(all,all)$}; ]
					\edge node [auto=left] {$data(401:800)$};
					[.\node (2) {$n=400$\\$Query(all,all)$}; ]]
				\edge node [auto=left] {$data(801:1600)$};	
				[.\node (6) {$n=800$\\$Query(list(4),list(5))$\\$Estimate D$};
					\edge node [auto=right] {$data(801:1200)$};
					[.\node (4) {$n=400$\\$Query(all,all)$}; ]
					\edge node [auto=left] {$data(1201:1600)$};
					[.\node (5) {$n=400$\\$Query(all,all)$}; ]]			
		]

	\tikzset{every node/.style={draw,rectangle,fill=white}}
	\foreach \x in {1,...,7} 
	 {
	 \node at (\x.north east) {\x};
	 };
\end{tikzpicture}
\caption{ActiveKmedoids workflow for 1600 points($b=2$,$t_h=400$)}
\label{fig:flowTree}
\end{figure}
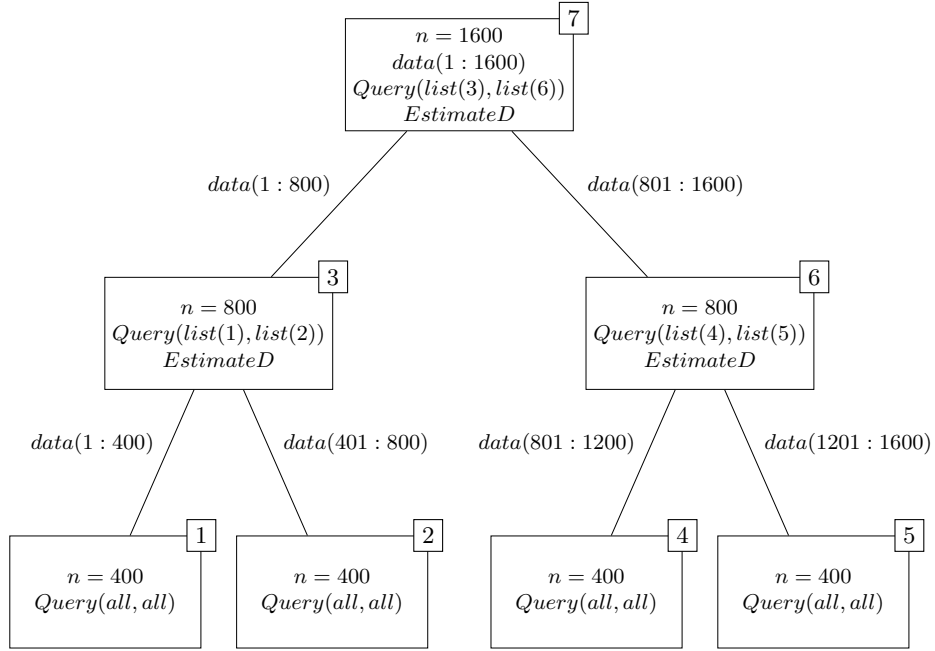

Figure~\ref{fig:flowTree} shows an example workflow for Algorithm~\ref{alg:red} for $1600$ data points with branching factor $2$ and threshold $400$. \par

Now we calculate the complexity of our Algorithm~\ref{alg:red}. 
It makes a tree with the branching factor $b$ and the threshold $t_h$ for the number of the data points in the leaves of the tree. Therefore, the height of this tree is $\ceil{\log_{b}{(n/t_h)}}$. 
The number of nodes in the $i$th level of the tree is $b^i$ and each node of the $i$th level has $n/b^i$ data points.
According to~\cite{kmd2011singh}, the time complexity of the k-medoids algorithm is $\BigO{kn^2}$ for $n$ data points and $k$ clusters in each iteration.
Consider $p$ as the maximum number of iterations used in the k-medoids algorithm, then the time complexity in each node in the $i$th level of the tree is $\BigO{(n^2/b^{2i})kp}$. 
Therefore, the overall complexity of Algorithm~\ref{alg:red} is
\begin{equation}
\BigO{\sum_{i=1}^{\floor{\log_{b}{(n/t_h)}}+1} (\frac{n^2}{b^{2i}}kp)b^i} = \BigO{n^2kp}.
\end{equation}

A major factor that measures the quality of an active clustering algorithm, is the number of distances that the algorithm demands from the oracle. The number of the asked queries in the internal nodes of the tree is $\BigO{b^2(k(s+1))^2}$ and in the leaves is $\BigO{nt_h}$. Since, in the proposed method, there are $n/t_h$ leaves and each of them has $t_h$ data points, the ratio of the asked distances to all of the distances is almost equal to
\begin{equation}
(b^2(k(s+1))^2\frac{b^{\floor{\log_{b}{(n/t_h)}}}-1}{b-1} + nt_h) / {n \choose 2}
\end{equation}
where $(b^{\floor{\log_{b}{(n/t_h)}}}-1)/(b-1)$ shows the approximate number of the internal nodes in the tree.

In order to improve the accuracy of the estimated distances, we can increase the $s$ value. However, it may raise running time of calculating upper-bound estimations and the number of queries that should be asked. 

To have a good baseline for comparison, we introduce an algorithm based on random selection of the distance queries, called \textit{Random-Rival} and compare results of our algorithm with those of this algorithm. In Section \ref{sec:experiments}, we show that asking queries using our method is much better than asking them randomly that is the aim of any active clustering algorithm. \par

Random-Rival(RR) algorithm, considers data points as the vertices of a weighted graph in which the weight of each edge shows the distance between its endpoints. Firstly, RR asks some distances randomly and then estimates all unknown distances using the triangle inequality and \textit{Floyd-Warshall}~\cite[p. 693]{clrs} algorithm. In the other words, we find shortest path distances of all pairs using the available distances in the aforementioned graph. Since our distance function is a distance metric, the length of these shortest paths is an upper-bound estimation of the true distances. Finally, RR clusters data runs the k-medoids algorithm on the estimated distances. Since Floyd-Warshall worst case time complexity is $\BigO{n^3}$~\cite[p. 695]{clrs}, we can consider the runtime complexity of RR as $\BigO{n^3}$ at worst. Algorithm~\ref{alg:RR} shows the pseudo code of the RR algorithm. \par

\begin{algorithm}
\caption{Random Rival}\label{alg:RR}
\begin{algorithmic}[1]
\INPUT $n,k,B$ \Comment{\#data, \#clusters, budget}
\OUTPUT $C_1,\ldots,C_k$ 
\Procedure{RandomRival}{$n,k,B$}

\State $D \gets n\times n$ infinity matrix
\State $(x_1,y_1),\ldots,(x_B,y_B) \gets $ random pairs such that $1 \leqslant x_{i} < y_{i} \leq n$
\State $\forall i, 1 \leqslant i \leqslant B:$ update $D(x_i,y_i)$ by querying distances
\State $D_e \gets FloydWarshall(D)$
\State $C_1,\ldots,C_k \gets kmedoids(D_e,k)$
\EndProcedure
\end{algorithmic}
\end{algorithm}

\section{Empirical results}
\label{sec:experiments}

In this section, we show the results of our algorithm on some synthesized and real world datasets.
General information about these datasets are presented in Table \ref{table:geninf}.
Most of them are real world datasets, but some are synthesized which are marked with letter s in Table \ref{table:geninf}. NORM-10 \cite{kmeanspp} contains $10000$ data points having $20$ features. This dataset has been generated by choosing $10$ real centers uniformly at random from the hypercube of side length $50$. Then, for each of the real centers, $1000$ points from a Gaussian distribution of variance one centered at the corresponding point is generated. We converted samples in NEC\_animal \cite{NEC2009deep} and ALOI200 \cite{aloi} dataset into $32\times 32$ grayscale images. ALOI \cite{aloi} (Object Viewpoint version) has $1000$ classes but we use only $200$ classes of it. \par

\begin{table}[]
\centering
\caption{General information about datasets}
\label{table:geninf}
\begin{tabular}{|l|c|c|c|c|c|}
\hline
Data           & \#Samples & \#Features & \#Classes & \#Distances    & Ref.               \\ \hline
vary-density(s)  & 150       & 2          & 3         & 11175           & \cite{ELKI}      \\ \hline
seeds          & 210       & 7          & 3         & 21945           & \cite{UCIrvine}    \\ \hline
mouse(s)       & 500       & 2          & 4         & 124750          & \cite{ELKI}     \\ \hline
fisheriris     & 150       & 4          & 3         & 11175           & \cite{UCIrvine}    \\ \hline
data\_2000(s)  & 2000      & 2          & 5         & 1999000         & \cite{data2000}    \\ \hline
Trace          & 200       & 275        & 4         & 19900           & \cite{UCR}        \\\hline
multi-features & 2000      & 649        & 10        & 1999000         & \cite{UCIrvine}    \\ \hline
TwoDiamonds(s) & 800       & 2          & 2         & 319600          & \cite{FCPS}    \\ \hline
EngyTime(s)    & 4096      & 2          & 2         & 8386560         & \cite{FCPS}    \\ \hline
COIL100        & 7200      & 1024       & 100       & 25916400        & \cite{columbia100} \\ \hline
NORM10(s)      & 10000     & 20         & 10        & 49995000        & \cite{kmeanspp}    \\ \hline
NEC\_animal    & 4371      & 1024       & 60        & 9550635         & \cite{NEC2009deep} \\ \hline
ALOI200        & 14400     & 1024       & 200       & 103672800       & \cite{aloi} \\ \hline
\end{tabular}
\end{table}

Although active version of some clustering algorithms like DBSCAN and spectral clustering have been introduced in  \cite{Mai2013actdbscan,wang2010}, these clustering algorithms are substantially different from the k-medoids algorithm. For example, DBSCAN and spectral clustering methods can find clusters of different shapes while k-medoids cannot.
Thus, we cannot compare results of our active k-medoid with those of active DBSCAN and active spectral clustering methods.
One way to evaluate an active clustering method that asks distances is to compare it with a clustering method that asks a random subset of distances. 
, we compare our method with the Random-Rival algorithm introduced in section \ref{sec:method} that tries to use a random subset of distances to estimate the whole distance matrix (using shortest paths on the graph of data points and the triangle inequality). It must be mentioned that in both the proposed active k-medoid and the Random-Rival algorithm, the clustering algorithm that is run on the obtained distance matrix will be k-medoids.
One of the most common measures for comparison of clustering algorithms is \textit{normalized mutual information} (NMI) \cite{Nguyen2009}. This measure shows the agreement of the two assignments, ignoring permutations. In the other words, NMI for the clustering obtained by an algorithm shows the agreement between the obtained grouping by this algorithm and the ground truth grouping of data. \par

We run our algorithm with $s=1,3$ and branching factor $b=2$ over all the datasets. Threshold $t_h$ is set to the minimum possible value which is equal to the number of classes for each dataset. Greater value of $s$, branching factor $b$, or threshold $t_h$ can improve NMI score for some datasets. However, it would also increase the number of queries which is usually quite unsatisfactory. We also run Random-Rival over these datasets which requires a specified proportion of distances starting from $5\%$ to $100\%$ (with the step $5$ percent). \par

Results of our method with the parameters mentioned in the previous paragraph are shown in Table \ref{table:resmeth}. For each algorithm and each dataset, NMI score and the ratio of the asked distances are presented in the table cells. For the Random-Rival algorithm, the results for the proportion of distances (between $5\%$ and $100\%$) that is the first place where the number of inquired distances is greater than or equal to the inquired ones in our method are reported. \par 

\begin{table}[]
\centering
\caption{NMI results of the methods. Numbers in parenthesis show percent of the asked distances.}
\label{table:resmeth}
\begin{tabular}{|l|c|c|c|}
\hline
Data           & RR            & s = 1         & s = 3          \\ \hline
vary-density   & 70.8 (10.0\%) & 95.0 (9.7\%)  & 96.6 (10.5\%) \\ \hline
seeds          & 55.7 (10.0\%) & 90.3 (7.1\%)  & 89.5 (7.3\%)  \\ \hline
mouse          & 58.1 (5.0\%)  & 75.5 (4.1\%)  & 73.6 (4.3\%)  \\ \hline
fisheriris     & 65.3 (10.0\%) & 85.6 (9.6\%)  & 89.3 (10.2\%) \\ \hline
data\_2000     & 88.5 (5.0\%)  & 77.1 (1.9\%)  & 78.3 (1.9\%)  \\ \hline
Trace          & 45.7 (10.0\%) & 51.2 (9.6\%)  & 52.4 (10.4\%) \\ \hline
multi-features & 46.9 (5.0\%)  & 78.7 (2.8\%)  & 77.6 (2.8\%)  \\ \hline
TwoDiamonds    & 97.5 (5.0\%)  & 100.0 (1.8\%) & 100.0 (1.8\%) \\ \hline
EngyTime       & 71.2 (5.0\%)  & 99.9 (1.6)    & 99.6 (1.6\%)  \\ \hline
COIL100        & 42.3 (10.0\%) & 75.6 (7.6 \%) & 75.6 (8.0\%)  \\ \hline
NORM10         & 95.5 (5.0\%)  & 94.3 (1.6\%)  & 94.6 (1.6\%)  \\ \hline
NEC\_animal    & 23.6 (10.0\%) & 66.6 (7.6\%)  & 67.3 (7.9\%)  \\ \hline
ALOI200        & 48.0 (10.0\%) & 79.6 (7.7\%)  & 79.7 (8.0\%)  \\ \hline
\end{tabular}
\end{table}

\begin{figure}[ht!]
\centering
\includegraphics[width=120mm]{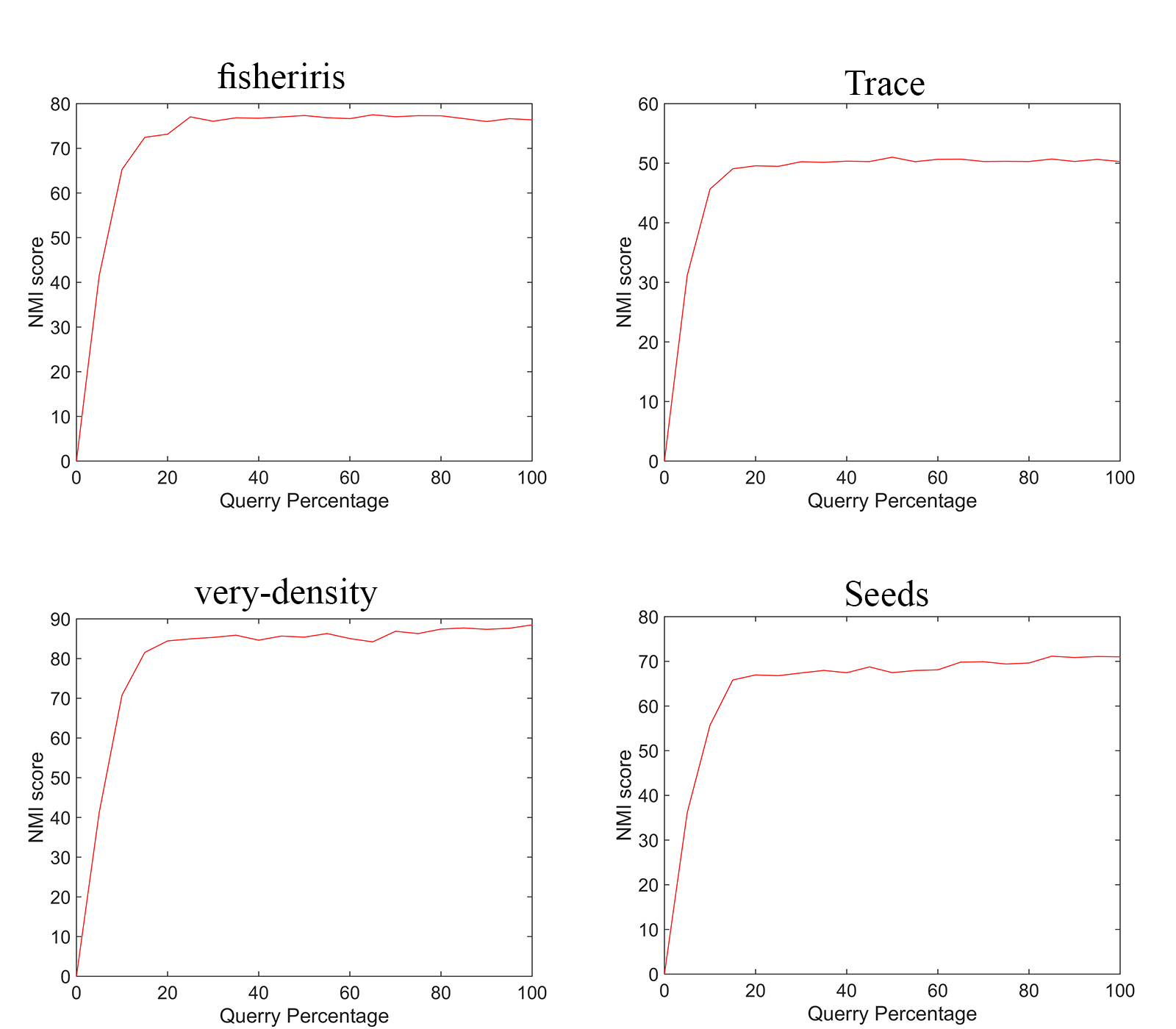}
\caption{Random-Rival over four datasets.
\label{fig:RR_Table}}
\end{figure}

Moreover, results of the Random-Rival algorithm which uses $0$ percent of distances up to $100\%$ of them is presented for some datasets in figure~\ref{fig:RR_Table}. According to figure~\ref{fig:RR_Table}, RR shows an ascending trend by asking extra distances and it gets close to the maximum value by asking about $20\%$ of distances. Although it sounds to be a good algorithm, the proposed active k-medoids algorithm gets better NMI values with asking fewer number of distances. Moreover, the time complexity of our active k-medoid is also better. These results state the power of our algorithms which find accurate clusters by asking only a small subset of distances. \par

\section{Conclusion}
\label{sec:conclusion}
In this paper, we introduce an innovative active distance-based clustering method. Its goal is to cluster $n$ points from a metric dataset into $k$ clusters by using lowest number of distances that is possible. We design a recursive model that makes a tree and split data with a branching factor $b$ unless the number of objects is less than a threshold $t_h$. Then, it actively selects and ask some pairwise similarities from and oracle. After that it tries to make an upper-bound estimation for unknown distances utilizing triangular inequality. Eventually, it clusters data with a simple k-medoids algorithm. We run our algorithm over some synthesized and real world datasets. In order to show privilege of our method and to compare the results, we also introduce an algorithm which randomly selects pairwise distances and estimates unknown ones using Floyd-Warshall algorithm.  \par


\small
\baselineskip=.85\baselineskip
\bibliographystyle{abbrv}
\bibliography{active_kmedoids}

\end{document}